\documentclass[letterpaper, 10 pt, conference]{ieeeconf}

\usepackage{amsmath,amsfonts}
\usepackage{array}
\usepackage[caption=false,font=normalsize,labelfont=sf,textfont=sf]{subfig}
\usepackage{textcomp}
\usepackage{stfloats}
\usepackage{url}
\usepackage{verbatim}
\usepackage{graphicx}
\usepackage{longtable}
\usepackage{booktabs}

\usepackage{algorithm}
\usepackage{algpseudocode}

\usepackage{cite}
\begin{document}

\title{Design Space Exploration of Low-Bit Quantized Neural Networks for Visual Place Recognition}

\author{Oliver Grainge, Michael Milford, Indu Bodala, Sarvapali D. Ramchurn and Shoaib Ehsan
\thanks{This work was supported by UK EPSRC Grants EP/Y009800/1 and EP/V00784X/1. Oliver Grainge, Indu Bodala, Sarvapali D. Ramchurn and Shoaib Ehsan are with the School of Electronics and Computer Science, University of Southampton, Hampshire, SO17 1BJ, United Kingdom (e-mail: oeg1n18@soton.ac.uk; i.p.bodala@soton.ac.uk; sdr1@soton.ac.uk; s.ehsan@soton.ac.uk). Shoaib is also with University of Essex, Colchester C04 3SQ, United Kingdom. Michael Milford is with the School of Electrical Engineering and Computer Science, Queensland University of Technology, Brisbane, QLD 4000, Australia (e-mail: michael.milford@qut.edu.au)}
\thanks{}}

\markboth{Journal of \LaTeX\ Class Files,~Vol.~14, No.~8, August~2021}%
{Shell \MakeLowercase{\textit{et al.}}: A Sample Article Using IEEEtran.cls for IEEE Journals}


\maketitle

\begin{abstract}
Visual Place Recognition (VPR) is a critical task for performing global re-localization in visual perception systems. It requires the ability to accurately recognize a previously visited location under variations such as illumination, occlusion, appearance and viewpoint. In the case of robotic systems and augmented reality, the target devices for deployment are battery powered edge devices. Therefore whilst the accuracy of VPR methods is important so too is memory consumption and latency. Recently new works have focused on the recall@1 metric as a performance measure with limited focus on resource utilization. This has resulted in methods that use deep learning models too large to deploy on low powered edge devices. We hypothesize that these large models are highly over-parameterized and can be optimized to satisfy the constraints of a low powered embedded system whilst maintaining high recall performance. Our work studies the impact of compact convolutional network architecture design in combination with full-precision and mixed-precision post-training quantization on VPR performance. Importantly we not only measure performance via the recall@1 score but also measure memory consumption and latency. We characterize the design implications on memory, latency and recall scores and provide a number of design recommendations for VPR systems under these resource limitations.

\end{abstract}


\section{Introduction}
Visual Place Recognition (VPR) is the task of recognising a previously visited place. It forms a critical part of the global re-localization module of visual simultaneous localization and mapping systems \cite{10054007}. VPR performs loop closures which reduce the uncertainty in perceptions \cite{Engel2014LSDSLAMLD}, improving the robustness of downstream tasks. To achieve these benefits a VPR system must be able to recognise when two images are taken from the same place. This seemingly simple task is challenged by natural variations that occur when images depicting the same place are taken at different times. These variations include illumination, occlusion, viewpoint and appearance changes. 

\begin{figure}\label{fig:block}
\centering
\vspace*{1ex}
\includegraphics[width=0.85\columnwidth]{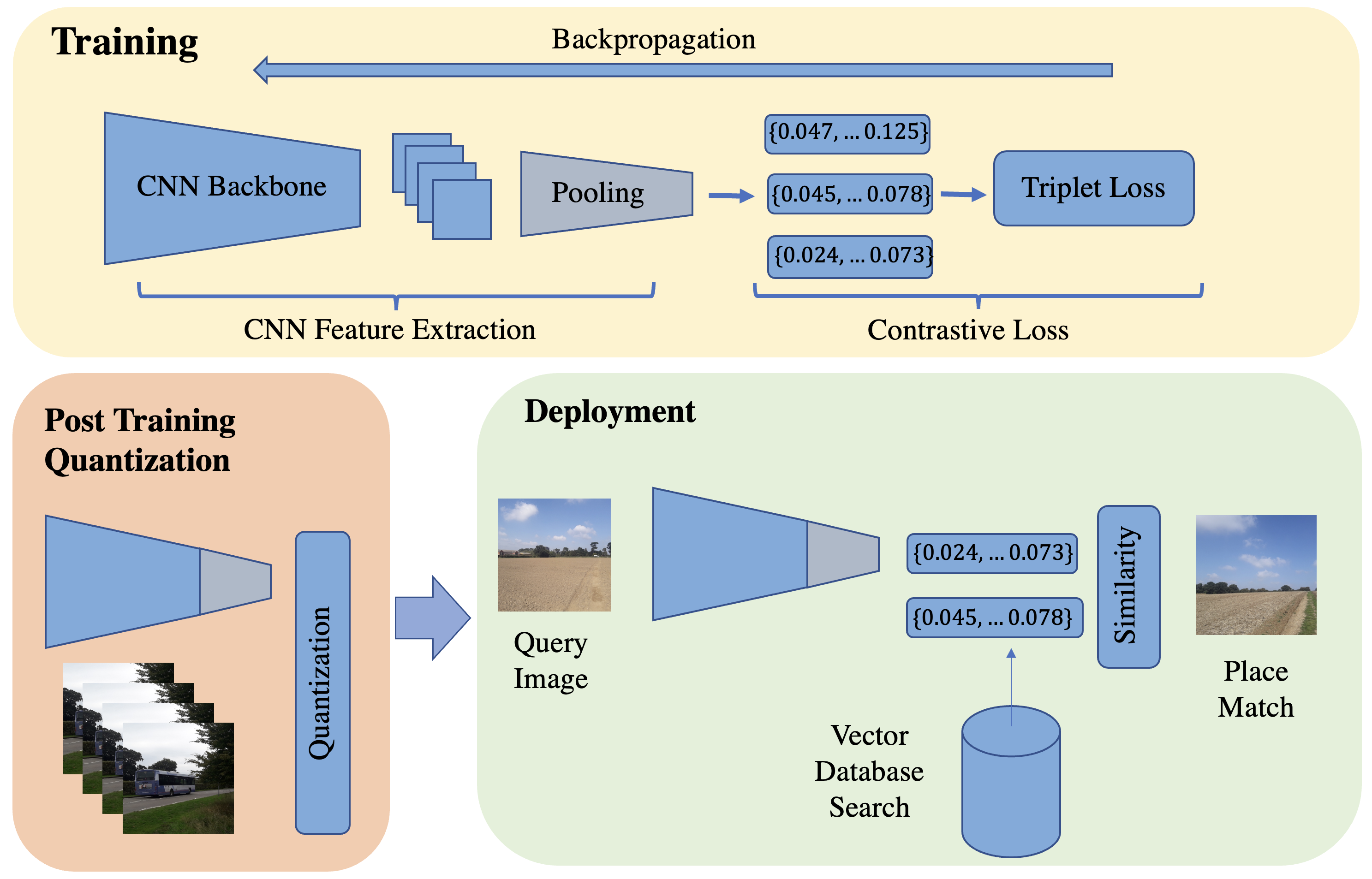}
\caption{Block Diagram of our Visual Place Recognition Training, Quantization and Deployment Pipeline}
\end{figure}

VPR is a representation problem. To achieve high performance a method must extract low dimensional features that are close in vector space for images of the same place whilst being distant for images of different places. A process depicted by the deployment section in Figure \ref{fig:block}. Previously this process has been achieved by lightweight handcrafted feature extraction techniques such as those utilizing keypoint descriptors \cite{sift, Bay2006, calonder2010brief} or whole image holistic descriptors \cite{gist, 6081878}. Handcrafted features, whilst highly efficient do not perform robustly under appearance or viewpoint changes \cite{zaffar2021vpr}. Robustness to appearance and viewpoint changes has been achieved through the design of specialized pooling layers in combination with deep convolutional neural networks \cite{spoc,  mac, netvlad, gem}. Whilst the convolutional backbone as shown in Figure \ref{fig:block} provides the capacity to learn respresentations through hierarchical layers of abstraction, the pooling layers shown in grey act on feature maps with different principles to obtain features with specific invariance properties. For instance GeM \cite{gem} utilizes a generalized mean layer to focus the activation distribution on salient image regions thus providing enhanced viewpoint invariance \cite{zaffar2021vpr}. NetVLAD on the other hand uses a differentiable relaxation of the VLAD descriptor improving invariance to appearance change \cite{zaffar2021vpr}. 

Recent work has shown that scaling of the network backbone and training dataset significantly improves recall scores \cite{gsvcities, cosplace}. The scaling however has resulted in the most performant models being infeasible for deployment on resource constrained embedded platforms such as those used by drones \cite{Hausler2021}. This is either because the model does not fit in the device memory or because it's limited throughput does not enable the system to satisfy it's real-time constraints.

In terms of constraints, Embedded systems, the primary target devices for VPR have limited memory, throughput, and power. However, with the proliferation of vision and machine learning applications, there is a growing trend toward enabling these systems to handle parallel computations, specifically with integer arithmetic, as evidenced by circuits such as systolic array compute units \cite{8425458}. Such advancements are driven by the benefits of reduced latency, energy, and memory consumption \cite{Wu_2016_CVPR}. They are realized through the reduced complexity of integer Multiply Add Accumulates (MAC's) and a reduction in the required memory bandwidth for reading and writing layer weights and activation's \cite{qq}. In essence quantized neural networks reduce resource utilization significantly.

In this work, we study the design of low-bit-width VPR networks in four dimensions: CNN
backbone, pooling method, quantization scheme, and descriptor size. We explore the interdependencies between these design points and hypothesize that by combining optimizations from each of these aspects, much of the recall performance of larger, more powerful networks can be preserved while making the computation tractable for resource-constrained platforms. Our quantization exploration includes both single and mixed-precision methods, with the latter's precision levels found through a genetic search. As such, our work identifies the specific design configurations required to maximize VPR recall performance under specific resource constraints, thus providing significant value to the VPR research and robotics communities.

\section{Related Work}
Forming image descriptors robust to appearance and viewpoint changes is an ongoing challenge for the VPR community. Early attempts used handcrafted features to identify key points. The Scale Invariant Feature Transform (SIFT) \cite{Lowe2004} identifies key points using the Difference Of Gaussians blob detector and describes them using a histogram of oriented gradients (HOG). SIFT descriptions are speeded up in SURF \cite{bay2006surf} by utilizing integral images. BRIEF \cite{calonder2010brief} a more lightweight keypoint descriptor produces binary features for efficient similarity search. Each of these keypoint feature description methods can be aggregated with a Bag Of Visual Words (BOVW), Fisher Vectors or Vector of Locally Aggregated Descriptors (VLAD) technique \cite{csurka2004visual, douze2011combining, arandjelovic2013all}. Whilst both VLAD and BOVW cluster keypoint features to form a codebook, BOVW computes its descriptor with a histogram of code frequencies, and VLAD the sum of residuals between features and their corresponding codes. Whilst effective, handcrafted features have struggled with viewpoint and appearance changes leading to the widespread use of CNN's \cite{zaffar2021vpr}. Excellent robustness to appearence change can be achieved with NetVLAD \cite{Arandjelovic16} an end-to-end trainable differentiable relaxation of the VLAD descriptor. NetVLAD has been extended in Patch-NetVLAD to process query image patches and perform a spatial consistency check during matching \cite{hausler2021patchnetvlad}. SPoC \cite{spoc} a simpler pooling method utilises spatial mean pooling and MAC \cite{mac} spatial max pooling. GeM \cite{gem} is a parameterized generalization between average and max pooling that can learn to focus its activation distribution on salient image regions.

Previous research in Visual Place Recognition (VPR), as highlighted in \cite{geobench}, has predominantly concentrated on recall performance, often overlooking memory and latency considerations. The potential of compact convolutional models, particularly those integrating grouped convolutions and inverted bottlenecks, remains underexplored in this domain. Quantization is a viable strategy for efficient VPR, focusing on minimizing the precision of weights and activations. In its most extreme form, binary neural networks represent weights and activations with a single bit, yielding impressive memory and latency efficiencies \cite{ferrarini2022highly, ferrarini2022binary}, thereby rendering them suitable for micro-controllers and low-power CPUs. Despite their advantages, these networks exhibit limitations in capturing the complex representations essential for large-scale VPR \cite{binnet}. While substantial progress has been made in integer quantization for classification and other vision tasks \cite{cnnquant1, cnnquant2}, its application to VPR remains largely unexplored.

\section{Experimental Setup}

\subsection{Datasets}

We train our models using the Mapillary Street Level Sequences Dataset \cite{msls}, a diverse collection of over 1.6 million images showing variations in illumination, viewpoint, appearance, and occlusions. For testing, the Nordlands, St Lucia, and Pittsburgh30k \cite{netvlad, nordland, stlucia} datasets are utilized where each assesses a different aspects of the VPR system. Nordlands offers seasonal appearance changes, Pittsburgh30k displays extreme viewpoint shifts, and St Lucia focuses on illumination variations. Together they validate VPR methods against a diverse set of challenges.

\begin{figure*}
    \centering
    \vspace*{1ex}
    \includegraphics[width=\textwidth]{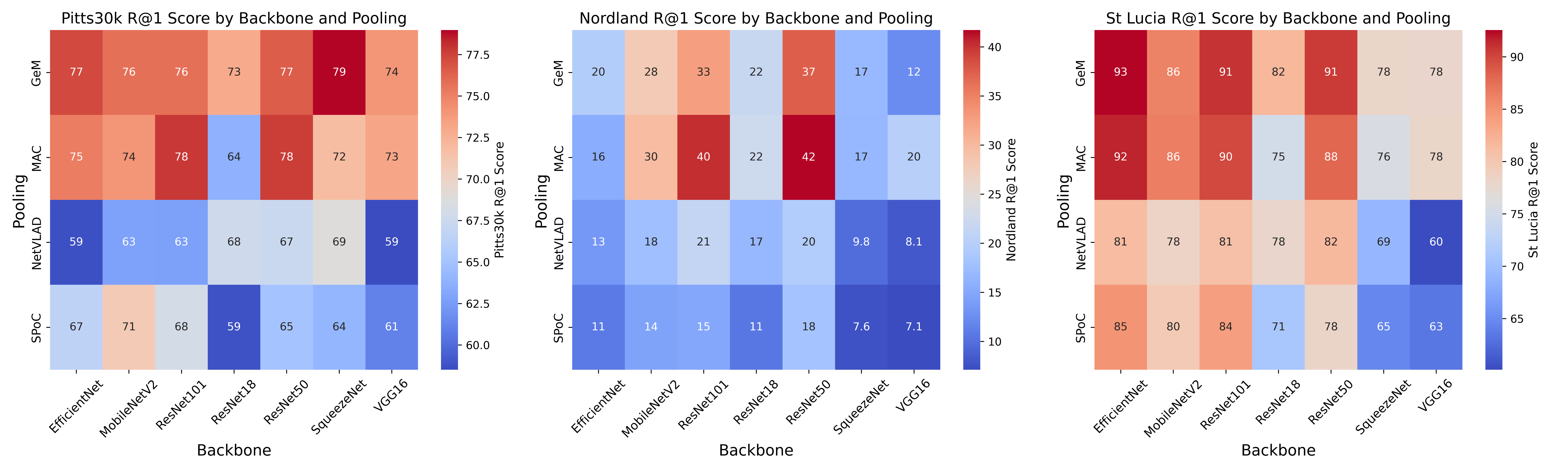}
    \caption{Network Recall@1 performance for different datasets, backbones, and pooling methods under fp16 quantization. As can be seen from the top two rows of the figure, the pooling method is most important for Recall@1 performance.}
    \label{fig:all_recalls}
\end{figure*}

\subsection{Training}
As illustrated in Figure \ref{fig:block}, all models are trained using contrastive learning, which uses the triplet loss function as defined in Equation \ref{eqn:contrastive}. This function reduces the distance between descriptors of the anchor images $a$ and positive images $p$ that depict the same location, using the cosine similarity function $f$. Concurrently, it increases the distance between the descriptors of the anchor images and negative images $n$. To enhance training efficiency, we mine the most challenging negatives from a subset of the dataset. We use the Adam optimizer with a learning rate of $1e^{-4}$ and stop training when validation accuracy hasn't increased for 3 epochs.

\begin{equation}\label{eqn:contrastive}
    L(a, p, n) = \max\left(0, f(a, p) - f(a, n) + \alpha\right)
\end{equation}

\subsection{Backbones}
In this work, we investigate the effect of the backbone convolution architecture on the embedded system performance for VPR. To this end, we test 7 different backbones including the traditional VGG-16, ResNet-18, ResNet-50 and ResNet-101 models \cite{Simonyan15, He_2016_CVPR}. The VGG-16 model provides a more traditional network using large kernel sizes without any batch normalization thus providing later layers with a large receptive field. ResNet's on  the other hand utilize deeper networks with a smaller kernel size making them parameter efficient. The use of ResNet's residual connections and batch normalization supports the propagation of gradients allowing for the design of deeper networks. We include different depths to investigate its effect on performance in terms of memory, latency and recall.

We additionally investigate the use of backbone architectures optimized for mobile inference including MobileNetV2, SqueezeNet and EfficientNet B0 \cite{Sandler2018, iandola2016SqueezeNet, Tan2019}. All three use factorized convolutions to increase parameter efficiency and inverted bottlenecks to maintain expression capacity. Inverted Bottlenecks expand the channel dimensions with 1x1 convolutions and subsequently manipulate information with efficient depth-wise convolution. MobileNetV2 \cite{Sandler2018} utilizes the inverted bottle-necks with a final linear projection to preserve information. SqueezeNet \cite{iandola2016SqueezeNet} uses Fire Modules, a version of the inverted bottleneck that utilizes both 1x1 and 3x3 depth-wise convolutions to capture more spatial features. EfficientNet \cite{Tan2019}, a model found by forming scaling laws for depth, resolution and width also uses a squeeze and excite mechanism to weight channels by their relative importance.

\subsection{Pooling}
The four pooling methods we investigated are NetVLAD, SPoC, GeM and MAC \cite{netvlad, spoc, gem, mac}. NetVLAD given a set of $d$ dimensional descriptors $X = \{x_0, x_1, \cdots, x_N\}$ and $K$ learnt codes $C = \{c_0, c_1, \cdots, c_K\}$ computes the weighted sum of residuals between the closest codes and descriptors as shown in Equation\ref{eqn:netvlad_}. The weights $a_{nk}$ are computed via a differentiable relaxation of the hard cluster assignment given by Equation \ref{eqn:netvlad_assign}.

\begin{equation}\label{eqn:netvlad_}
v_k = \sum_{n=1}^{N} a_{nk} (x_n - c_k)
\end{equation}

\begin{equation}\label{eqn:netvlad_assign}
a_{nk} = \frac{\exp(\langle x_n, c_k \rangle)}{\sum_{j=1}^{K} \exp(\langle x_n, c_j \rangle)}
\end{equation}

SPoC \cite{spoc} is a spatial weighted average pooling. Given a convolutional feature map of dimensions $D$x$H$x$W$. The spatial dimensions $H$x$W$ are averaged producing a $D$ dimensional descriptor. In this work, we adopt uniform weighting making SPoC akin to global average pooling. The MAC \cite{mac} descriptor is very similar and implemented as global max pooling. We also investigate GeM which is a learnable generalization between SPoC and MAC \cite{spoc, gem}. As shown in Equation \ref{eqn:gem} given $D$ feature maps $F$ with spatial dimensions $H$x$W$ the generalization is performed with the learnable parameters $p$.

\begin{equation}\label{eqn:gem}
v_d = \left( \frac{1}{HW} \sum_{i=1}^{H} \sum_{j=1}^{W} F_{ijd}^p \right)^{\frac{1}{p}}
\end{equation}

\subsection{Quantization}
As shown in Figure \ref{fig:block} we perform post-training quantization (PTQ) using a data sample to reduce memory, energy consumption and minimize latency. PTQ to floating point 16 (fp16) requires only a type cast. As fp16 has a 5-bit exponent and 10-bit mantissa as opposed to fp32's 8-bit exponent and 23-bit mantissa it has a significantly smaller dynamic range and numeric precision.

\begin{figure}
\centering
\vspace*{1ex}
\includegraphics[width=\columnwidth]{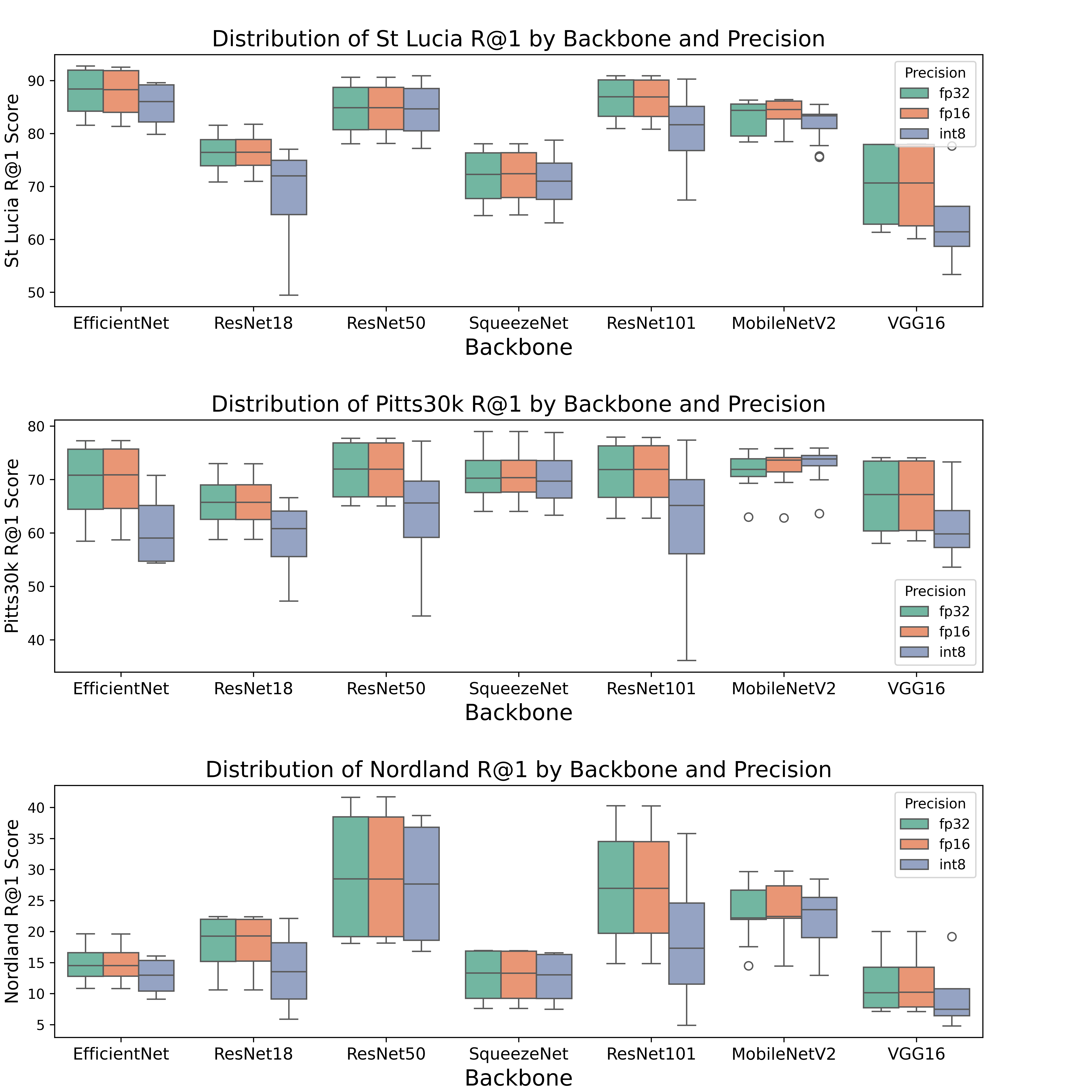}
\caption{Distribution of convolutional backbone recall@1 performance under full-network quantization. The figure shows no loss in performance under fp16 precision.}
\label{fig:backbone_performance}
\end{figure}

Quantizing the network for integer inference is more involved as continuous values can no longer be represented. In this work, we quantize weights to a signed 8-bit or 4-bit integer data type for weights and accumulate in int32 for activations. We utilize linear symmetric quantization over linear asymmetric as it removes an additional zero point bias and improves implementation efficiency in hardware. The quantization is performed by Equation \ref{eqn:quantize} with $b$ the integer bit width and $s$ the floating point scaler. Dequantization is performed by Equation \ref{eqn:dequantize}. For convolution weights, we use per-channel granularity with a scaler $s$ for every channel. For activations, per-tensor granularity is used with just a single scale parameter. Per-channel granularity for weights has been chosen as the factorized convolutions used by MobileNetV2, SqueezeNet and EfficientNet \cite{Sandler2018, iandola2016SqueezeNet, Tan2019} have significantly varying weight distributions across channels \cite{qmobilenet}. To reduce the information loss under quantization we use entropy calibration by selecting a scale $s$ that minimizes the Kullbeck-Leibler divergence between the quantized and unquantized weights.

\begin{equation}\label{eqn:quantize}
    w_{int} = \text{clamp} (\lfloor \frac{w_{float}}{s} \rceil, 0, 2^{b-1}) 
\end{equation}

\begin{equation}\label{eqn:dequantize}
    w_{float} = s \circ w_{int}
\end{equation}

To further improve numerical stability and reduce latency we perform layer fusion for all quantization levels. For networks which utilize convolution followed by batch normalization, we fuse the the layers together. Given a convolutional filter $W \in \mathbb{R}^{c, h, w, c'}$ and activation $X \in \mathbb{R}^{n,h,w,c}$ the convolution can be described by Equation \ref{eqn:convolution}. A batch-normalization is given by Equation \ref{eqn:batchnorm}. Fusing them together requires the substitution of Equation \ref{eqn:convolution} into \ref{eqn:batchnorm} giving Equation \ref{eqn:layer_fusion}. The fusion eliminates an additional read and write from memory and reduces quantization noise. 

\begin{equation}\label{eqn:convolution}
    X' = W * X + b
\end{equation}

\begin{equation}\label{eqn:batchnorm}
    Y_{i,j,k,c'} = \lambda_{c'} \frac{X'_{i,j,k,c'} - \mu_{c'}}{\sqrt{\sigma^2_{c'} + \epsilon}} + \beta_{c'}
\end{equation}

\begin{equation}\label{eqn:layer_fusion}
    Y = \frac{\lambda}{\sqrt{\sigma^2 + \epsilon}} W X + \beta + \frac{\lambda}{\sqrt{\sigma^2 + \epsilon}}(b - \mu)
\end{equation}

\section{Architectural Performance}
In this section, we evaluate the performance impact of compact architectural design. For each design, we evaluate the Recall@1 accuracy in addition to the memory and latency cost. All evaluations are performed with a compact 1024 dimension descriptor using the TensorRT back-end on a single consumer GPU.

\subsection{Backbone}\label{sec:backbone}
\begin{figure}
\centering
\vspace*{1ex}
\includegraphics[width=\columnwidth]{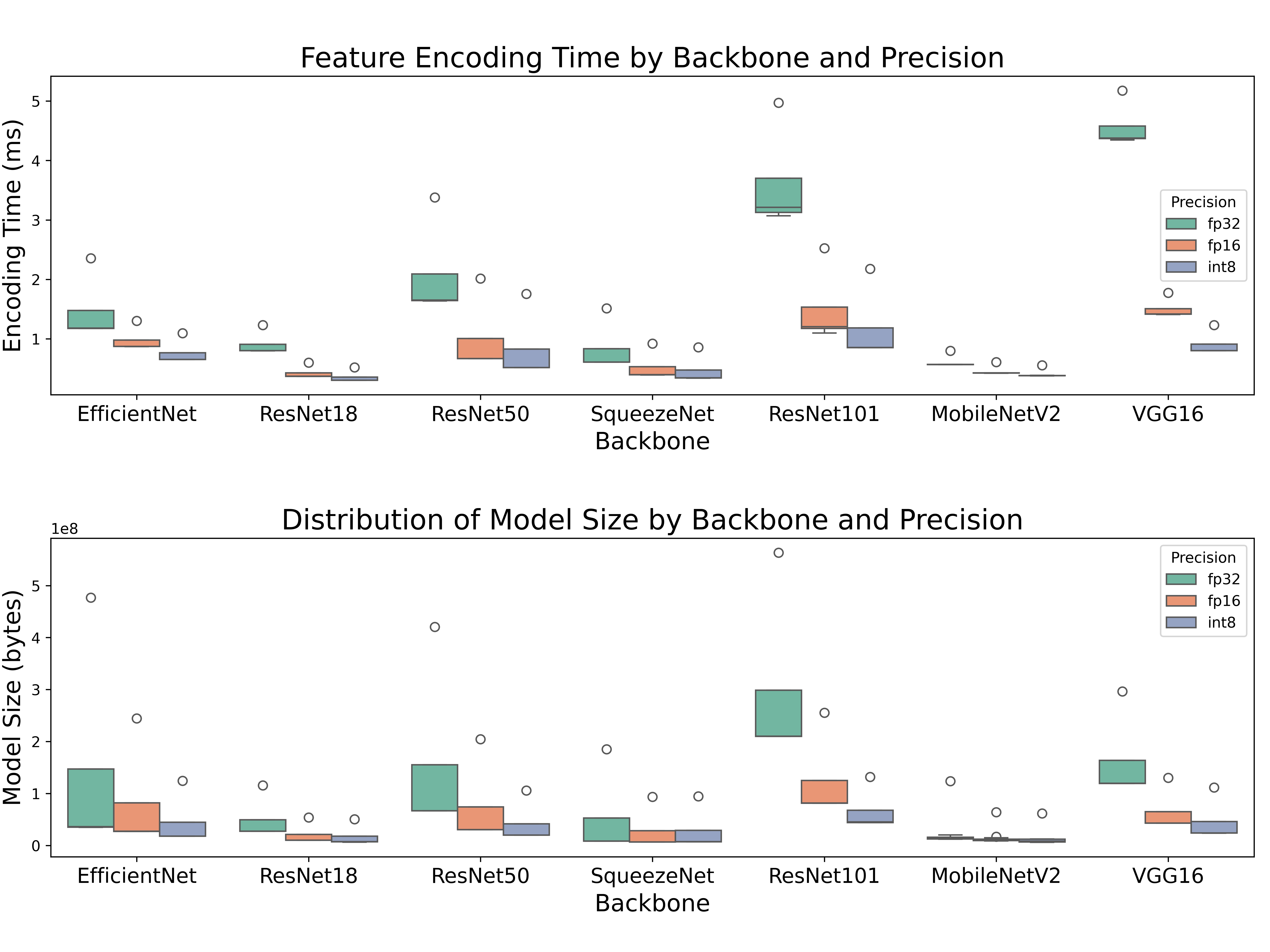}
\caption{Latency and memory cost for convolutional backbone architectures (outliers are caused by the NetVLAD models significant resource consumption)}
\label{fig:backbone_size_latency}
\end{figure}

Figure \ref{fig:backbone_performance} displays backbone recall performance across datasets, pooling layers, and quantization precisions. Notably, quantizing to fp16 preserves recall performance for all architectures, while halving memory and significantly reducing latency (Figure \ref{fig:backbone_size_latency}). However, PTQ to int8 often drops recall@1 scores due to quantization noise. The most robust backbones under int8 precision are MobileNetV2 and SqueezeNet, with MobileNetV2 showing superior recall generalization.

 EfficientNet excels on the St Lucia dataset but underperforms with Nordlands' appearance changes. It shares MobileNetV2's inverted bottlenecks but lacks a non-linear activation at the end of its convolution block. Its Squeeze-and-Excite mechanism, using global average pooling for channel weighting, may be less critical for VPR.

ResNet-50, a larger network, matches MobileNetV2's performance. Extending ResNet beyond 50 layers, as shown in Figure \ref{fig:backbone_performance}, doesn't improve recall but increases latency and memory usage. While there are benefits to expanding beyond 18 layers, lightweight models with depthwise-separable inverted bottlenecks are more suited for VPR whilst optimizing latency and memory.

\begin{figure}
\centering
\vspace*{1ex}
\includegraphics[width=\columnwidth]{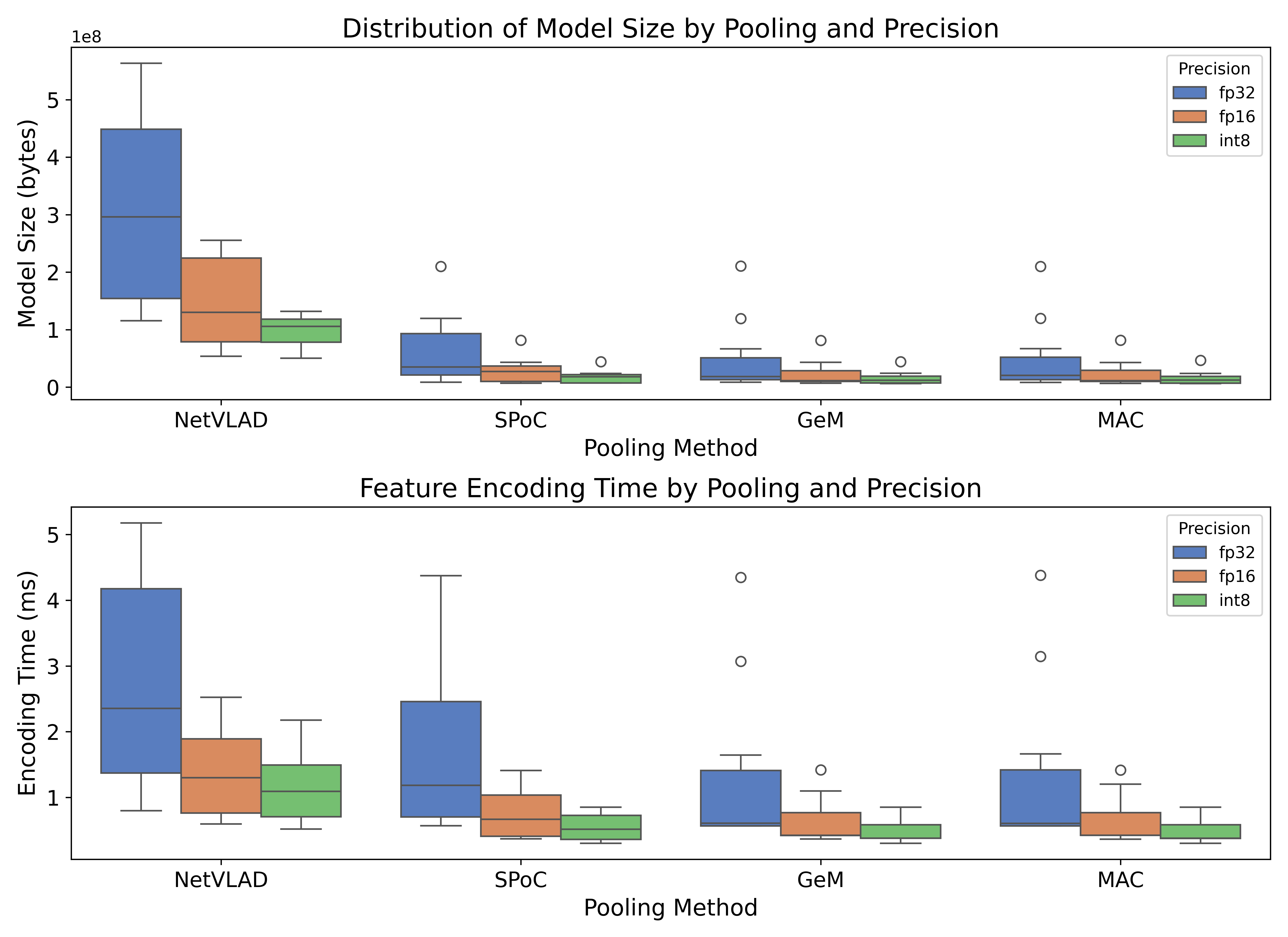}
\caption{Distribution of model latency and memory utilization under different backbones, precisions and pooling techniques. MAC and GeM pooling show significantly reduced latency and memory consumption.}
\label{fig:aggregation_size_latency}
\end{figure}

\subsection{Pooling Method}\label{sec:aggregation}
\begin{figure}[!t]
\centering
\vspace*{1ex}
\includegraphics[width=\columnwidth]{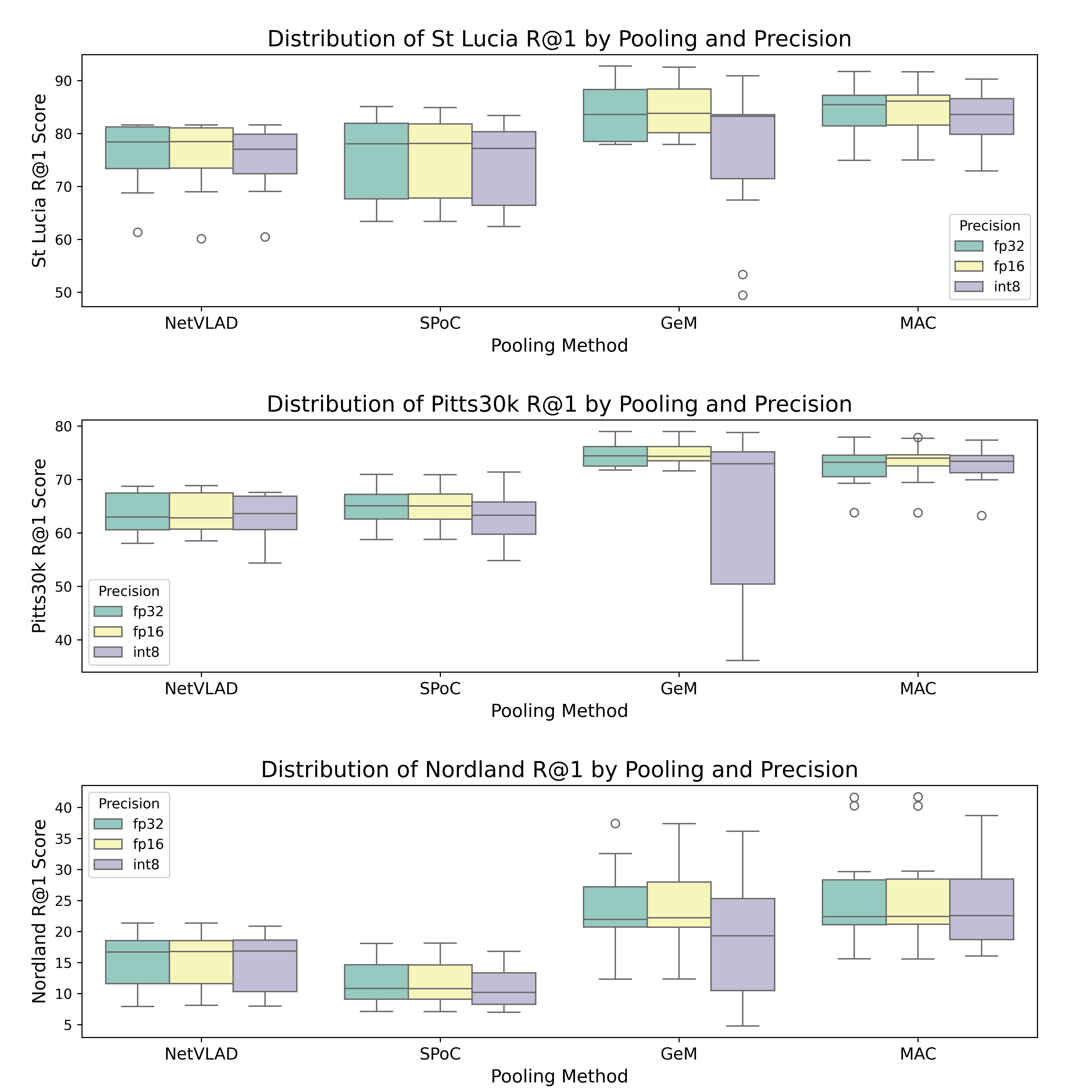}
\caption{Distribution of pooling layer recall@1 performance under full network quantization to different precisions. GeM and MAC pooling shows consistently higher recalls.}
\label{fig:aggregation_perf}
\end{figure}

As shown by Figure \ref{fig:aggregation_perf} the most performant pooling methods in terms of recall@1 across all three test datasets are GeM and MAC. Whilst GeM shows slightly higher recall@1 on the St Lucia and Pittsburgh datasets in both fp32 and fp16 precisions it's performance significantly drops during quantization to int8. This is due to the sensitivity of the activation's given by Equation \ref{eqn:gem} to the quantization noise in parameters $p$. As such if int8 quantization is required for memory and latency improvements MAC pooling should be chosen over GeM, SPoC or NetVLAD.

Interestingly as shown by Figure \ref{fig:aggregation_size_latency} the NetVLAD pooling has a significant impact on the model latency and memory consumption. This is because the model has to store an additional set of parameters in the NetVLAD codes $C$, as shown by Equation \ref{eqn:netvlad_}. The activation descriptor also has a very high dimension equal to the number of codes multiplied by the code dimension. In this work, we use 64 codes with dimension equal to the number of channels in the activation of the last convolutional feature. The additional fully connected linear layer following the NetVLAD descriptor projects the representation to a size of 1024 and hence consumes a significant amount of memory. 

SPoC and MAC pooling methods are parameterless pooling layers and hence have no impact on memory size. They also have a negligible impact on encoding time in comparison to the backbone encoding time. GeM has a very small number of parameters, just one per convolutional feature map and hence provides little impact on the memory size of the model and encoding time. 

\section{Descriptor Size}
In Sections \ref{sec:backbone}, \ref{sec:aggregation}, and Figure \ref{fig:all_recalls}, we found that the MobileNetV2 backbone with either MAC or GeM pooling offers the best memory, latency, and recall trade-off. Therefore, we focus on this design subset to study the effect of varying descriptor size. Larger descriptors improve VPR feature discrimination but increase memory use and retrieval latency. Here, we explore how descriptor size affects encoding and retrieval latency, and recall performance.

\begin{table*}[ht]
\centering
\vspace*{2ex}
\caption{Impact of Descriptor Size on recall performance and latency. All results are computed at floating point 16 precision.}
\begin{tabular}{l|r|r|r|r|r|r|r}
\toprule
\text{backbone+pooling} & \text{Descriptor Size} & \text{Pitts30k R@1} & \text{Nordland R@1} & \text{St Lucia R@1} & \( \tau_r \) & \( \tau_e \) & \( \tau_{\text{total}} \) \\
\hline
\midrule
\text{MobileNetV2+GeM} & 512 & 71.5962 & 21.8614 & 78.5519 & \textbf{0.0485} & \textbf{0.4220} & \textbf{0.4705} \\
\text{MobileNetV2+GeM} & 1024 & 72.1244 & 22.2238 & 86.2022 & 0.0908 & 0.4248 & 0.5156 \\
\text{MobileNetV2+GeM} & 2048 & 74.3251 & 27.9574 & 83.8115 & 0.1720 & 0.4245 & 0.5965 \\
\text{MobileNetV2+GeM} & 4096 & \textbf{75.7776} & 26.5657 & 83.8115 & 0.3396 & 0.4272 & 0.7668 \\
\hline
\text{MobileNetV2+MAC} & 512 & 69.4249 & 22.3362 & 85.1776 & 0.0490 & 0.4243 & 0.4733 \\
\text{MobileNetV2+MAC} & 1024 & 73.2688 & 27.1637 & 85.5191 & 0.0904 & 0.4240 & 0.5144 \\
\text{MobileNetV2+MAC} & 2048 & 73.9730 & 22.4449 & 86.1339 & 0.1720 & 0.4258 & 0.5978 \\
\text{MobileNetV2+MAC} & 4096 & 74.0317 & \textbf{29.7333} & \textbf{86.4071} & 0.3410 & 0.4262 & 0.7672 \\
\hline
\bottomrule
\end{tabular}
\label{tab:descriptor_dim}
\end{table*}

Table \ref{tab:descriptor_dim} shows that using the MobileNetV2 backbone with either MAC or GeM pooling on the Pitts30k dataset results in an average recall@1 increase of 4\% when the descriptor size is expanded from 512 to 4096. However, this increase in recall@1 score varies across datasets: it is only 1.2\% on St Lucia, but reaches 7.4\% on Nordland. This variation shows the importance of a larger descriptor size for VPR under appearance changes. On average, enlarging the descriptor size from 512 to 4096 leads to a recall@1 score improvement of 7.48\% with MAC pooling and 8.22\% with GeM pooling.

Increased dimensions however significantly impact latency and memory. We assessed the retrieval latency impact using an optimal brute force inner product search, revealing $O(N)$ encoding latency complexity with map size $N$ and $O(D)$ with descriptor size $D$. The table indicates that the retrieval latency $\tau_r$ measured under a 1k map size significantly affects total VPR system latency $\tau_{total}$, the sum of encoding $\tau_e$ and retrieval $\tau_r$ latency. Specifically, a descriptor size jump from 512 to 4096 raises VPR system latency by $63\%$ and memory use by 8x. This substantial increase in resource utilization highlights the importance of small descriptors for efficient VPR.

\section{Mixed Precision}

While integer post-training quantization significantly reduces memory consumption by an average of 69.49\% and latency by 62.68\%, it also leads to a decrease in recall performance by an average of 6.87\%. With that in mind, a balance can be struck between preserving recall performance and reducing precision on a per-layer basis. A search can be performed to identify the optimal per-layer precision that maximizes recall subject to an average bit-width budget $B$. As the vast search space makes a brute force search intractable we utilize a genetic algorithm (GA) as shown in Algorithm \ref{alg:ga} to efficiently search the configuration space.

\begin{figure}
\centering
\includegraphics[width=\columnwidth]{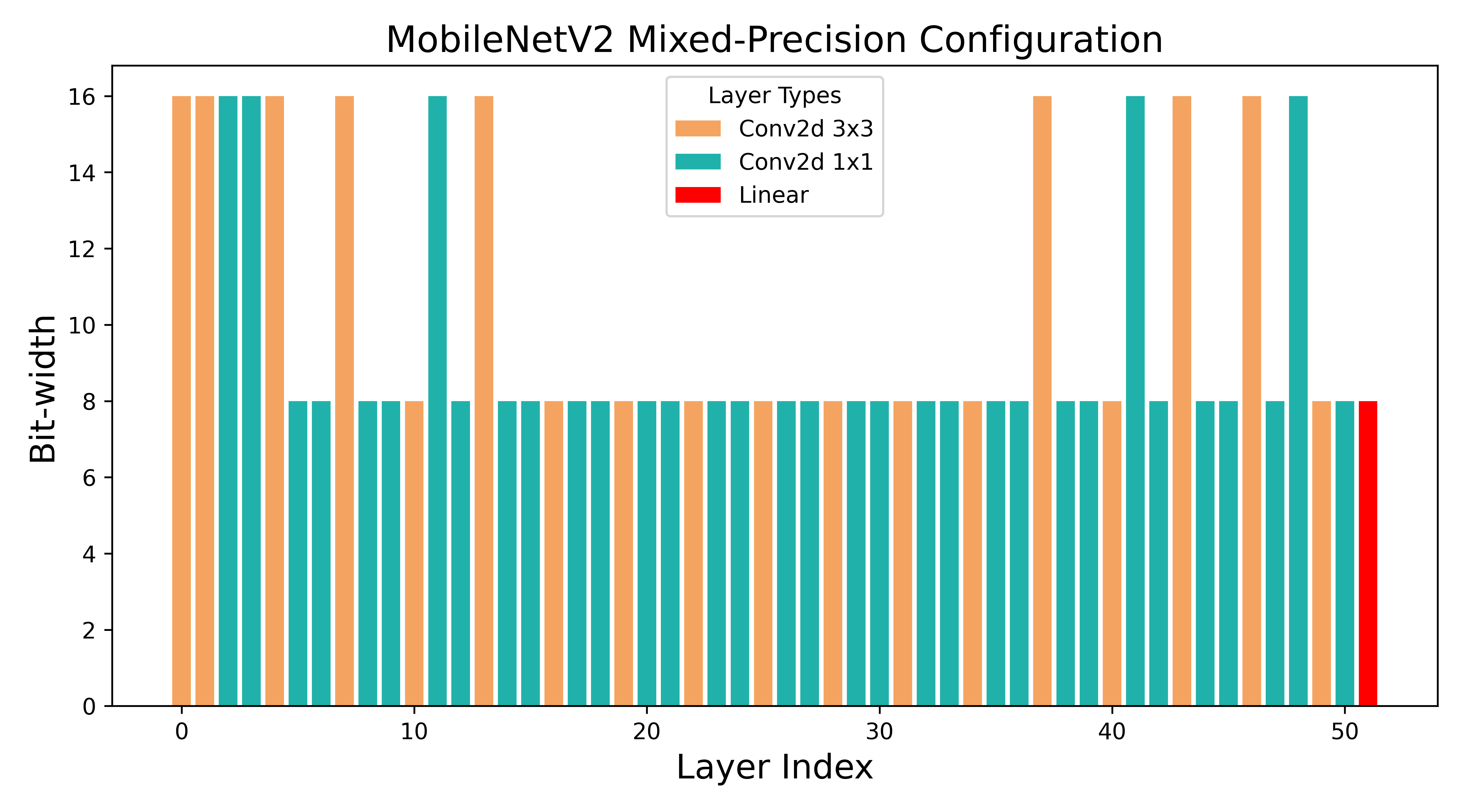}
\caption{Mixed-Precision network configuration found under evolutionary search with bit-width budget of 10. All batch-norm layers have been folded into the previous convolution}
\label{fig:mixed_config}
\end{figure}

  The GA's population $P$ is represented by $N$ mixed-precision candidates $I$ in which $I_i$ represents the bit precision of layer $i$. We limit the available precisions to int4, int8 and fp16 as those are the precision's currently supported for acceleration in commodity hardware. Given a full precision model $M$ and model $Q$ quantized under configuration $I$ we define the fitness function in Equation \ref{eqn:fitness}. Given a limited data sample $\mathbf{x}$ the objective function favours quantization configurations that minimize the difference between the full precision model descriptors and the quantized model descriptors. Whilst just being a proxy for recall accuracy, its evaluation is an order of magnitude faster than evaluating recall on the full test dataset, allowing for such a search to be completed. As in \cite{EvoQ} our Mutation function randomly alters the precision of a random layer with a probability $p_m$, where precisions are sampled from a categorical distribution skewed by the sensitivity of the layer to quantization noise. We however additionally utilize crossover between the two most fit individuals from the population by concatenating random sections of the two parents as we find it improves convergence.

\begin{equation}
\begin{aligned}\label{eqn:fitness}
& \underset{x}{\text{maxamize}}
& & f(I) = -\frac{1}{L}\sum_{i=0}^L (Q(I; x_i) - M(x_i))^2 \\
& \text{subject to}
& & B \geq \frac{1}{T} \sum^T_{i=0} I_i  \quad & \\
\end{aligned}
\end{equation}

Figure \ref{fig:mixed_config} shows a mixed-precision bit-width configuration found by the evolutionary algorithm with a bit-width budget of $10$. It can be seen that higher precisions are favoured in the earlier layers of the network. This may be because the early convolution layers that perform low-level pattern recognition and feature extraction are sensitive to noise. Additionally, quantization of early layers would inject noise into the network that may be amplified by later layers creating a larger objective loss of Equation \ref{eqn:fitness}. It is also seen empirically that some higher precision convolution layers are also required to preserve performance later in the network. If using a final linear layer for projection, a lower precision is adequate due to a high level of redundancy in its dense connections.

\begin{figure}
\centering
\includegraphics[width=\columnwidth]{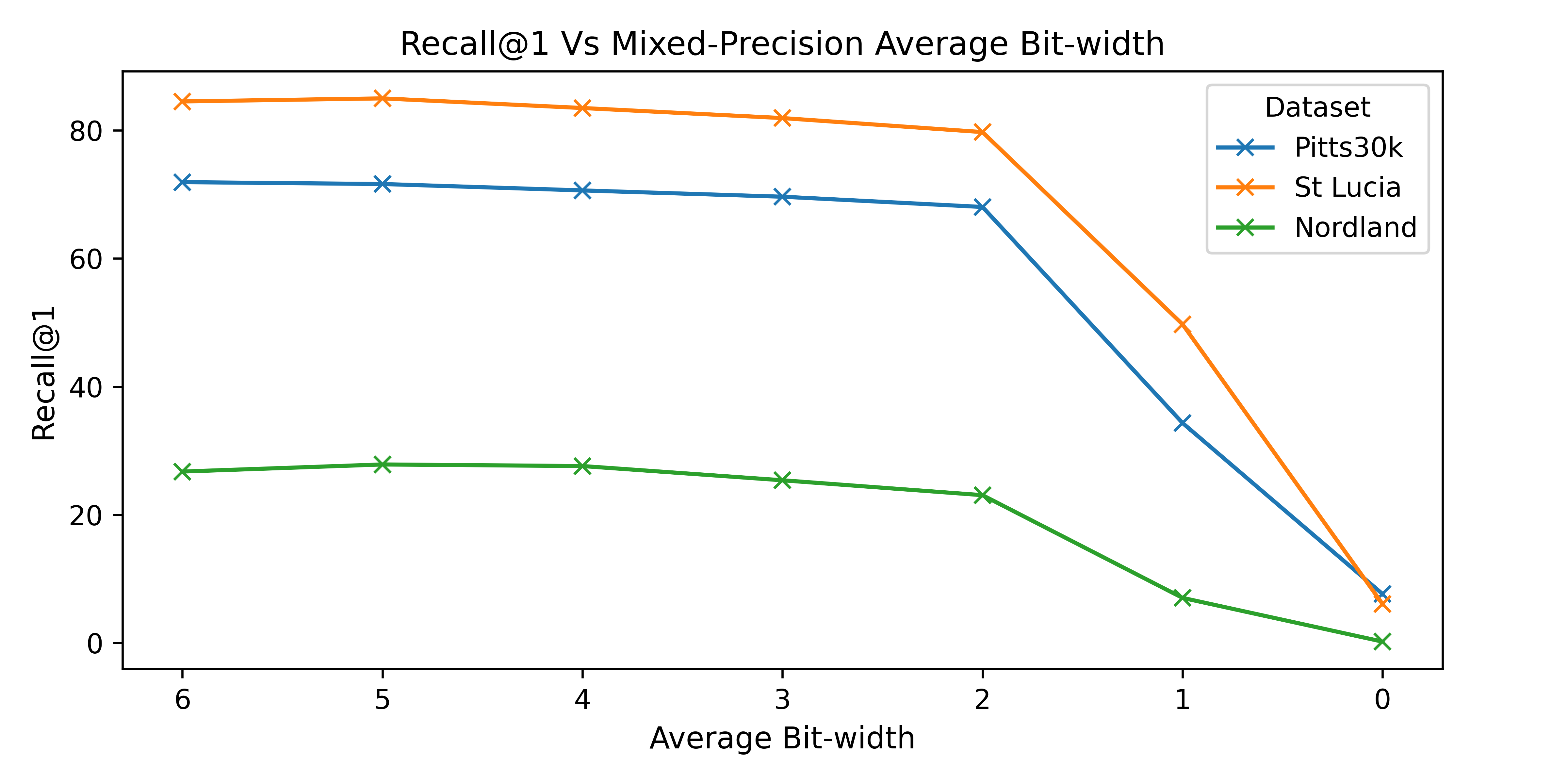}
\caption{Recall@1 performance under Mixed-Precision network configurations found with an evolutionary search. }
\label{fig:bitwidth}
\end{figure}

In Figure \ref{fig:bitwidth} we additionally investigate the trade-off between the recall and the average bit-width across the layers of the network $\bar{I}$. Whilst reducing the average bit-width decreases the latency and memory footprint of the network it also results in a reduction in recall score. A noticeable trend seen across datasets is the significant drop in performance when imposing an average bit-width budget $B$ from Equation \ref{eqn:fitness} below a value of 10.

\begin{algorithm}
\caption{Genetic Algorithm Mixed-Precision Search}
\label{alg:ga}
\begin{algorithmic}[1]
    \State \textbf{Input:} Objective function \( f(I) \), Population size \( N \), Mutation rate \( p_m \), Sample size \( C \), Average Bitwidth \(B\)
    \State \textbf{Output:} Best solution found $I^*$
    \State 
    \State Initialize population \( P \) with \( N \) random mixed-precision candidate solutions
    \While{termination condition not met}
        \State Sample \( C \) individuals from \( P \) to form \( P^s \)
        \State Select parents \( x_1, x_2 \) from \( P^s\) based on fitness \( f \)
        \State Remove worst individual from \( P^s \) based on \( f \)
        \State \( offspring \gets \) Crossover(\( x_1, x_2 \))
        \State \( offspring \gets \) Mutate(\( offspring \), \( p_m \)) subject to \( B \)
        \State Push \( offspring \) into \( P \)
    \EndWhile
    \State \Return the solution in \( P \) with the highest \( f(I) \)
\end{algorithmic}
\end{algorithm}

\section{Design Recommendations and Discussion}
Our work has provided an extensive investigation of the effectiveness of VPR performance optimizations in 4 dimensions. Through this experimentation, we have uncovered a number of valuable insights underpinning the best design practices for building advanced VPR systems under specific resource constraints. The most significant are as follows;

\textbf{Pooling}: The choice of pooling method has the largest impact on the generalization performance of the visual place recognition model. GeM and MAC show superior performance with a lower dimensional descriptor in terms of recall, memory consumption and latency. At floating point precision GeM shows a small recall improvement over MAC and hence should be chosen. At integer precisions however, It's sensitivity to noise causes significant performance drops. At this precision, MAC should be chosen over GeM.

\textbf{Backbone}: The backbone choice is more flexible allowing for a selection that meets the required latency and memory limits of the target device. A ResNet50 backbone provides excellent VPR generalization performance at the expense of 97Mb of memory. Should this exceed device limits MobileNetV2 performs well with just 14Mb of memory and shows significant reductions in latency. If this still does not meet the requirements of the embedded system low-bit quantization should be favoured over the use of smaller networks.

\begin{equation}\label{eqn:mem_constraint}
    M_{memory} > N \cdot D \cdot 4
\end{equation}

\textbf{Quantization}: Quantizing the model weights to fp16 halves the memory footprint and reduces the feature encoding time on average by 55\%. Hence fp16 should always be used in a deployment scenario provided there is hardware support. If further memory reductions are required below the 6.8Mb of half-precision MobileNetV2, int8 quantization should be used. However, if this degrades performance unsatisfactorily and there is still some headroom in terms of memory and latency, a balance can be struck with mixed-precision quantization. The mixed-precision quantization configuration for VPR should leave the initial convolutional layers in higher precision and perform low-bit quantization of the intermediate backbone layers. The final layers excluding any dense fully connected layers should also be kept at higher precision. This configuration however should always be subject to the constraint that the average layer bit-width of the network is above 10 otherwise significant recall performance will be lost. 

\begin{equation}\label{eqn:descriptor_dim}
    D = \frac{T - k_2 \cdot N} {k_1} + \tau_e
\end{equation}

\textbf{Descriptor dimension}: The largest dimension descriptor should always be chosen. However, as the dimension affects memory consumption it limits the size of the map that can be stored. It also significantly affects the total VPR system latency $\tau_{total}$ by increasing the retrieval time $\tau_r$. During inference, the VPR system must first extract features from a query image, and subsequently use them to search for a place match in the map. Provided the VPR system optimally searches all features in the database its complexity is linear with descriptor dimension and database size. Thus given the scale of the environment in terms of the number of map images $N$ and the latency target $T$, the descriptor dimension should be chosen according to Equation \ref{eqn:descriptor_dim} where $k_1$ is the time taken per unit increase in descriptor dimension and $k_2$ is the per unit time taken per increase in the number of map images $N$. This selection must be made subject to the constraints in \ref{eqn:mem_constraint} where $M_{memory}$ is the memory allocated for the database stored in $fp32$.

\section{Conclusion}
Our study investigates four aspects of visual place recognition (VPR) optimization: backbone architecture, pooling methods, descriptor dimensions, and quantization schemes. The findings illuminate key strategies for crafting efficient VPR systems, tailored to specific resource constraints. Essential insights include the impact of pooling on model generalization, the strategic selection of backbone architecture based on device capabilities, and the delicate balance between quantization and performance. These guidelines aid in developing VPR algorithms that are not only effective but also mindful of operational constraints, ensuring optimal performance within given resource limitations. This contributes significantly to the advancement of VPR technology, optimizing computational efficiency and recognition accuracy.

\bibliographystyle{IEEEtran}
\bibliography{references}

\newpage

\vfill

\end{document}